\documentclass[a4paper,fleqn]{cas-sc}

\usepackage[authoryear,longnamesfirst]{natbib}

\def\tsc#1{\csdef{#1}{\textsc{\lowercase{#1}}\xspace}}
\tsc{WGM}
\tsc{QE}

\usepackage{amsmath}
\usepackage{longtable}
\usepackage{tabularx}
\usepackage{booktabs}

\begin{document}
\let\WriteBookmarks\relax
\def\floatpagepagefraction{1}
\def\textpagefraction{.001}

\shorttitle{Tool-Integrated Collaborative Learning in STEM}    

\shortauthors{C. Fuster-Barceló et al.}  

\title [mode = title]{Scaffolding Collaborative Learning in STEM: A Two-Year Evaluation of a Tool-Integrated Project-Based Methodology}  



%


\author[1]{Caterina Fuster-Barceló}[orcid=0000-0002-4784-6957]
\credit{Conceptualization, Methodology, Validation, Formal analysis, Investigation, Writing – Original Draft, Visualization}

\author[1]{ Gonzalo R. Ríos-Muñoz}
\credit{Writing – Review \& Editing}

\author[1]{ Arrate Muñoz-Barrutia}
\credit{Conceptualization, Writing – Review \& Editing, Supervision, Project Administration, Resources} 

\cormark[3]
\cortext[3]{Corresponding author}

\ead{mamunozb@ing.uc3m.es} 



\affiliation[1]{organization={Biongineering Department},
            addressline={Universidad Carlos III de Madrid}, 
            city={Leganés},
            postcode={28911}, 
            state={Madrid},
            country={Spain}}






\begin{abstract}
This study examines the integration of digital collaborative tools and structured peer evaluation in the Machine Learning for Health master’s program, through the redesign of a Biomedical Image Processing course over two academic years. The pedagogical framework combines real-time programming with Google Colab, experiment tracking and reporting via Weights \& Biases, and rubric-guided peer assessment to foster student engagement, transparency, and fair evaluation. Compared to a pre-intervention cohort, the two implementation years showed increased grade dispersion and higher entropy in final project scores, suggesting improved differentiation and fairness in assessment. The survey results further indicate greater student engagement with the subject and their own learning process. These findings highlight the potential of integrating tool-supported collaboration and structured evaluation mechanisms to enhance both learning outcomes and equity in STEM education.
\end{abstract}




\maketitle









\section{Introduction}\label{sec:intro}

The rapid adoption of digital technologies and collaborative learning models is transforming higher education, offering new ways to address the needs of diverse and evolving learning environments. This shift accelerated during the COVID-19 pandemic, which exposed both the potential and the limitations of online education, particularly the challenge of sustaining student engagement and translating collaborative work into virtual formats \cite{ALON2023104870, GHERGHEL2023104795, SIDI2023104831, TSAI2023104849}.

To meet these challenges, teaching approaches must be adaptable, resilient, and capable of leveraging technology to support interaction, teamwork, and shared problem-solving. Computer-Supported Collaborative Learning (CSCL) provides a robust framework for this purpose, using digital platforms to counteract the isolation of remote learning and to strengthen learning communities \cite{MUNOZCARRIL2021104310}.

In STEM education, tools such as Weights \& Biases (W\&B)\footnote{\url{https://wandb.ai/site}} and Google Colab\footnote{\url{https://colab.research.google.com}} enable authentic, project-based learning experiences that mirror professional workflows. They offer shared cloud environments for co-developing code, tracking experiments, and generating reproducible outputs—practices that build technical expertise, critical thinking, and collaboration skills \cite{caeiro2021teaching, volkov2022using}. The increased use of hybrid and online learning further underscores the importance of frameworks that actively promote participation and collaboration across both physical and virtual settings \cite{alwafi2023impact}.

\begin{figure}
    \centering
    \includegraphics[width=\textwidth]{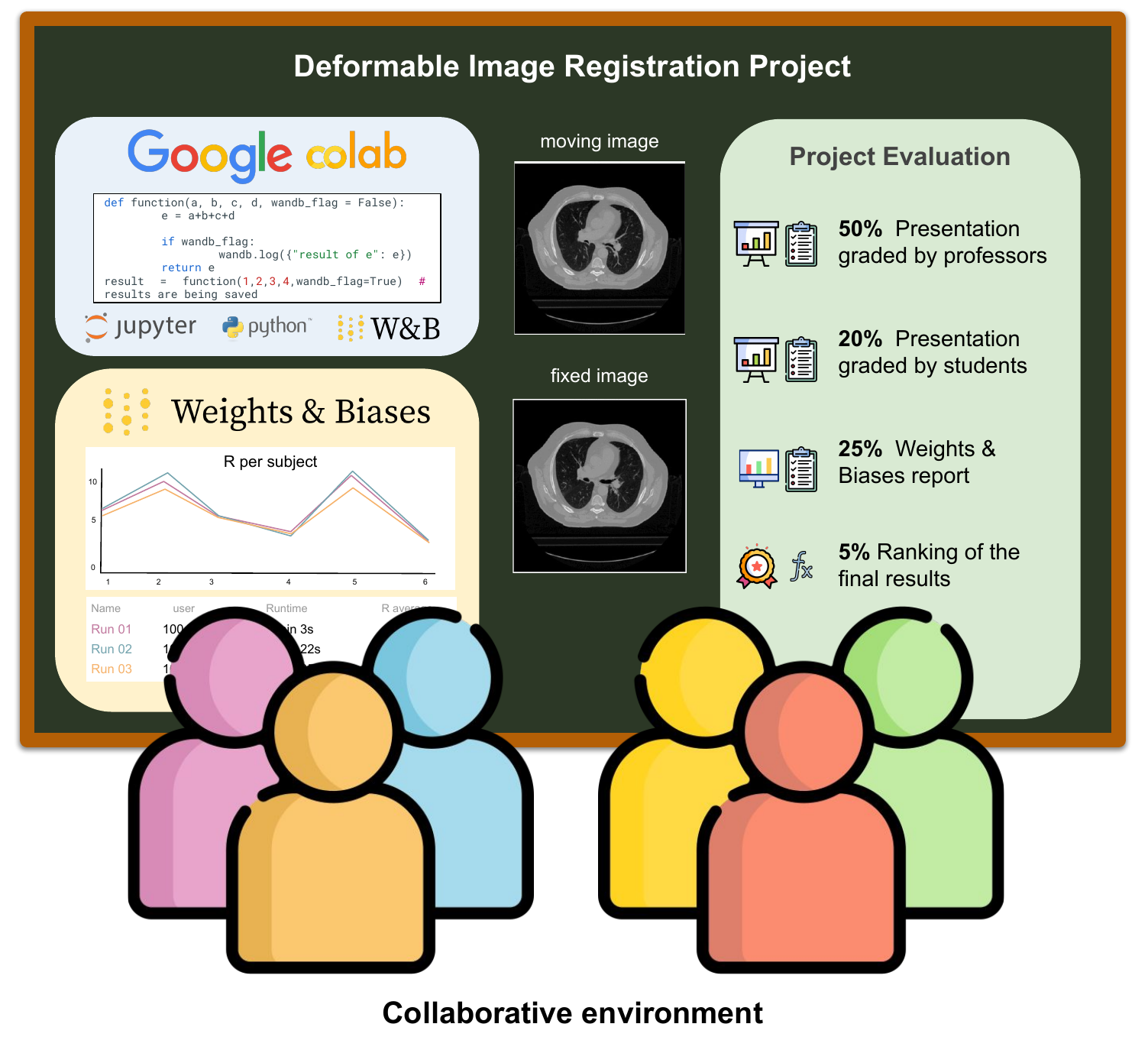}
    \caption{Graphical abstract illustrating the collaborative project-based methodology applied to the Deformable Image Registration task. The integration of Google Colab and Weights \& Biases enabled reproducible experimentation and visualization. The evaluation strategy incorporates educator and peer grading, performance ranking, and structured reporting to ensure fair and multifaceted assessment.}
    \label{fig:graphical-abstract}
\end{figure}

This study investigates the integration of W\&B, Google Colab, and structured peer evaluation in a master-level Biomedical Image Processing course (Figure \ref{fig:graphical-abstract}). The CSCL-based framework combines rubric-guided peer and instructor assessment with collaborative project work to ensure transparency, accountability, and reproducibility \cite{haleem2022understanding}. The goal is to enhance technical proficiency and collaborative competence while ensuring fairness and sustained student engagement.

By analyzing two consecutive cohorts, the study provides a longitudinal assessment of the approach’s scalability and consistency. The results contribute empirical evidence on the value of integrating collaborative digital tools into engineering and health technology education.

\section{Literature Review}

The transformative potential of digital tools and collaborative learning in education has been increasingly recognized, reflecting a shift toward more engaging, flexible, and interactive pedagogical practices. Traditional educational models, characterized by passive learning and limited student interaction, are being re-evaluated in light of digital innovations that offer enhanced learning experiences across disciplines \cite{dopson2010monologue}. The integration of technologies such as Artificial Intelligence (AI) and Machine Learning (ML) tools has marked a significant evolution in educational practices (\cite{zhai2021review, tahiru2021ai}). These technologies not only enrich learning experiences across a wide array of disciplines but also introduce a level of dynamism and inclusivity previously unattainable in conventional settings. 

Furthermore, the social constructivist theory emphasizes that learning is inherently social and shaped by cultural context, highlighting the value of collaborative learning environments \cite{vygotsky1987collected}. Empirical evidence supports this view, showing that students learn more effectively when they actively construct knowledge through peer interaction and cooperation \cite{MUNOZCARRIL2021104310}.

In the domain of CSCL, digital platforms have facilitated unprecedented levels of interaction among students, enabling collective problem-solving and knowledge construction. The efficacy of CSCL tools in increasing student engagement and achievement underscores the critical role of technology in fostering more dynamic and participatory learning experiences \cite{hernandez2019computer}.

The use of ML tools in education extends beyond theory, fostering practical, industry-relevant skills that are essential for today's workforce (\cite{9474408}). Innovations such as AI-based chatbots for responsible teaching, assistive educational robots, and courses on ML Operations (MLOps)\footnote{MLOps: \url{https://ml-ops.org}} illustrate how real-world applications can be integrated into academic curricula (\cite{LEE2022104646, PAPADOPOULOS2020103924, lanubile2023teaching}). This approach not only equips students to meet professional demands but also highlights the versatility and value of ML skills across sectors, from finance (\cite{xu2022mlops})  to manufacturing (\cite{mateo2023industry}) and beyond (\cite{subramanya2022devops}). 

Digital assessment tools have brought greater objectivity and efficiency to evaluating student work. By streamlining grading and revealing learning patterns, these tools enable personalized feedback and support, enhancing the overall educational experience \cite{simsek2021making, cosi2020formative}.

In conclusion, the integration of digital tools and collaborative learning strategies represents a pivotal shift in educational practices, offering opportunities to improve learning outcomes, student engagement, and the development of critical skills for the digital age. The broad potential of these approaches highlights the importance of continued research and innovation in this evolving field.

\section{Methodology}\label{sec:meth}

This manuscript describes the development and refinement of an innovative pedagogical approach in a Master-level engineering education. The approach combines advanced technological tools, including Python and other collaborative platforms, with traditional teaching methods to foster a hands-on, team-based learning environment. Its two-year implementation offers a strong foundation for assessing its impact on both technical skills and collaborative competencies.

\subsection{Pilot study}

The pilot study that informed the methodology was conducted in the 2022/2023 academic year within the 
\emph{AI for Health} course of the UC3M Master in Applied Artificial Intelligence.
In this course, students developed ML pipelines to analyze genomic and microbiological datasets. They applied techniques such as feature extraction and predictive modeling to address real-world problems, including antibiotic resistance prediction and cancer subtype classification. The aim was to build competencies in reproducible, data-driven decision-making using industry-standard tools. 

In the pilot, the evaluation focused mainly on model performance, measured by the final metric achieved by the neural networks, and a group presentation. However, the absence of a structured rubric or evaluation matrix reduced grading objectivity and consistency. This highlighted the need for a transparent, criteria-based assessment model, later developed in the Biomedical Image Processing course.

W\&B was systematically integrated as a collaborative platform for experiment tracking, results visualization, and producing group reports. Beyond its technical role, W\&B served as a shared workspace that supported core CSCL principles, including mutual visibility of progress, distributed responsibility, and peer-to-peer engagement. Its use also extended to structured weekly reports, simulating industry-style project reporting and strengthening transferable skills.

The pilot showed that W\&B enhanced both the transparency of group work and the creation of reproducible workflows, which motivated its formal adoption in later editions of the \emph{Biomedical Image Processing} course. In this manuscript, the pilot serves as a proof of concept, providing the foundation for a longitudinal study on the impact of CSCL-aligned digital tools on collaborative learning and skill development in biomedical engineering education.

\subsection{Participants and Course Context}

This study was conducted within the \emph{Biomedical Image Processing} course, part of the 
UC3M Master's in Machine Learning for Health. 
The course is delivered in person and combines theoretical instruction with project-based learning. Across two consecutive academic years (2023/2024 and 2024/2025), a total of 49 full-time students (25 for the first year and 24 for the second) participated in the pedagogical intervention described in this work. All students had prior experience in programming and foundational ML, acquired through earlier coursework in the program.

The course project, which represents 30\% of the subject's final grade, required students to work in teams of two to three on the development of a deformable image registration pipeline applied to thoracic CT (Computed Tomography) scans captured in different respiratory phases. Implemented in Python using Google Colab and tracked via W\&B, the project aimed to align medical images while promoting good practices in collaborative coding, reproducible experimentation, and peer feedback. The topic was chosen for its technical complexity, clinical relevance, and suitability for structured experimentation.

Although the final deliverable was submitted at the end of the term, students were encouraged to consult with instructors through open tutorships, particularly in the last two weeks. Two in-class project work sessions were also offered: one at the beginning of the project and another closer to the deadline. Attendance in the second session was notably low; only a few students participated, likely due to competing deadlines in other courses that were scheduled earlier, prompting students to prioritize those tasks.

Throughout both years, the instructional approach emphasized the integration of the CSCL principles. Students were expected to coordinate their work asynchronously using shared Colab notebooks and W\&B dashboards, with full transparency of individual contributions and experiment histories. Two faculty members supervised the course, providing guidance and formative feedback on request. Structured rubrics and evaluation criteria were shared in advance to clarify expectations and support self-regulated learning.

\subsection{Tools Integration for Collaborative Learning: W\&B and Google Colab}

Throughout both academic years, the course adopted a technology-enhanced learning setup that deliberately aligned with the principles of CSCL. The combined use of W\&B and Google Colab formed the backbone of this approach, enabling synchronous and asynchronous collaboration, transparent workload distribution, and structured reflection on experimental outcomes.

Google Colab served as the primary development environment, providing a cloud-based platform for real-time coding where students could jointly implement and test their registration pipelines. Built-in sharing and commenting features supported continuous code co-development and troubleshooting within teams.

Complementing this, W\&B was used to record each experiment, track performance metrics, and document model configurations. Each run was automatically associated with the responsible student or team, allowing instructors and peers to monitor progress and compare strategies. The ability to visualize metrics over time, annotate results, and store outputs in a centralized dashboard promoted transparency and encouraged knowledge exchange across teams.

This dual-tool setup also streamlined final project submissions. Students prepared their reports using W\&B’s integrated reporting tools, embedding logged metrics, annotated plots, and visualizations directly from their experiments. This workflow reinforced reproducibility and familiarized students with practices common in both academic research and industry projects.

\begin{figure}
    \centering
    \includegraphics[width=\textwidth]{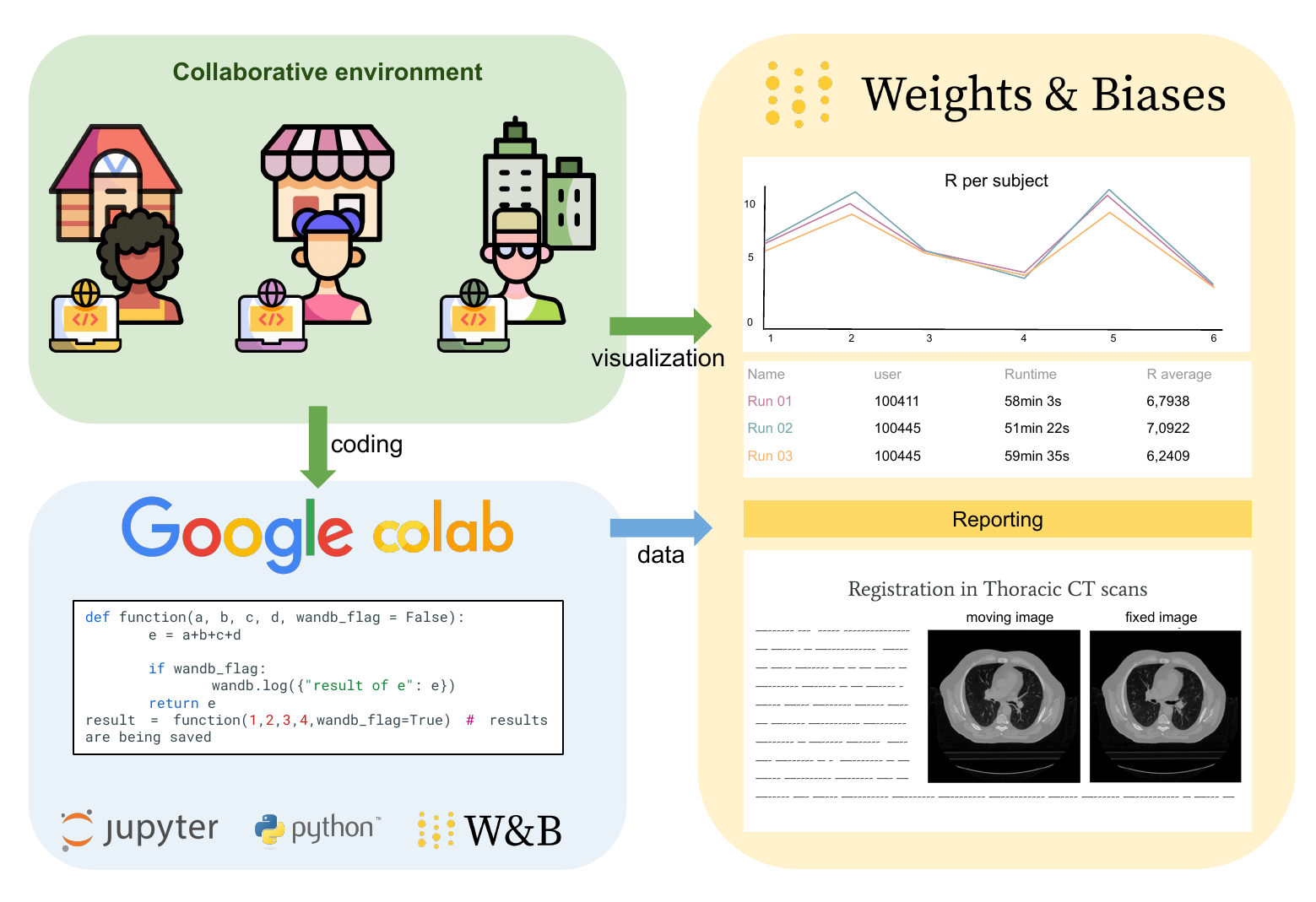}
    \caption{Integration of Collaborative Tools in Biomedical Image Processing. This figure illustrates the dynamic educational ecosystem combining Google Colab's interactive coding platform with W\&B for enhanced project tracking and reporting. The workflow showcases how students utilize Google Colab for code execution and W\&B for real-time visualization and report generation, culminating in an effective collaborative learning environment to solve the given task, in this case, the registration of thoracic CT scans.}
    \label{fig:wandb-and-googlecolab}
\end{figure}

Figure~\ref{fig:wandb-and-googlecolab} illustrates how these tools were embedded in the project workflow. By design, the setup advanced key pedagogical objectives: fostering collaborative knowledge construction, ensuring contribution transparency, and enabling formative feedback; central elements of CSCL-based instruction. 

\begin{table}[ht]
\centering
\caption{Collaborative features supported by the use of W\&B and Google Colab}
\label{tab:collaborative_affordances}
\renewcommand{\arraystretch}{1.3}
\begin{tabularx}{\textwidth}{>{\raggedright\arraybackslash}p{2.5cm} 
                                    >{\raggedright\arraybackslash}X 
                                    >{\raggedright\arraybackslash}X 
                                    >{\raggedright\arraybackslash}p{3.2cm}}
\toprule
\textbf{Tool} & \textbf{Feature} & \textbf{Pedagogical Function} & \textbf{CSCL Principle} \\
\midrule
Google Colab & Real-time co-development of code & Supports synchronous collaboration and immediate peer feedback within teams & Joint task engagement; Shared problem space \\
W\&B & Transparent experiment tracking per user & Allows monitoring of individual contributions and progress & Mutual awareness \\
W\&B & Visual comparison of registration results across teams & Facilitates inter-group knowledge sharing and reflective practice & Social comparison; Distributed cognition \\
W\&B & Collaborative report authoring with embedded metrics and visualizations & Reinforces reproducibility and joint interpretation of results & Shared knowledge construction; Externalization \\
W\&B + Instructor access & Live dashboards for monitoring and formative feedback & Enables just-in-time guidance and scaffolding by instructors & Guided participation; Monitoring and regulation \\
\bottomrule
\end{tabularx}
\end{table}

To ensure that the digital tools supported the course’s learning objectives, we deliberately matched features of Google Colab and W\&B with the needs of collaborative group work. Table~\ref{tab:collaborative_affordances} summarizes how these tools were used to support core functions of CSCL, such as shared task engagement, individual accountability, and mutual awareness. This structured integration ensured that students were not only co-developing code and reports but also engaging in transparent, trackable, and pedagogically meaningful interactions throughout the project lifecycle.

\subsection{Evaluation Framework}\label{subsec:evaluation}
The evaluation strategy was designed to holistically assess not only students’ technical performance but also their collaboration, engagement in peer feedback, and communication skills; key elements of the CSCL model that guide this course.

\subsubsection{Overview and Weighting}

The project evaluation was designed to balance technical rigor with collaboration, aligning with the course's learning outcomes. Table~\ref{tab:evaluation-breakdown} outlines the four assessment components, their pedagogical purpose, and their weight in the final project grade. 

\begin{table}[ht]
\centering
\caption{Breakdown of project evaluation components and their intended pedagogical goals.}
\label{tab:evaluation-breakdown}
\renewcommand{\arraystretch}{1.3}
\begin{tabularx}{\textwidth}{>{\raggedright\arraybackslash}p{3cm} 
                            >{\raggedright\arraybackslash}X 
                            >{\raggedright\arraybackslash}p{2.5cm}}
\toprule
\textbf{Component} & \textbf{Description and Pedagogical Intent} & \textbf{Weight} \\
\midrule
Project Report & Assess the team’s ability to document, analyze, and justify their registration pipeline with clarity and reproducibility & 25\% \\
Best Registration Result & Introduces a quantitative benchmark for experimental success while encouraging fair comparison across teams & 5\% \\
Oral Presentation & Evaluates individual student understanding, articulation, and communication of project contributions & 50\% \\
Peer Evaluation & Promotes critical engagement with peers’ work and fosters accountability and reflection within teams & 20\% \\
\bottomrule
\end{tabularx}
\end{table}

\subsubsection{Assessment Components and Rubrics}
The assessment was structured around the course’s core pedagogical objectives: developing reproducible research practices, maintaining technical rigor, and fostering collaborative learning. To address these aims, several complementary components were included, each targeting specific cognitive and interpersonal skills—from data-driven reasoning to reflective peer evaluation.

\paragraph{Project Report Rubric.} 
The project report was evaluated with a focus on three educational goals: clarity in scientific documentation, critical analysis of experimental design, and thoughtful engagement with the complexities of biomedical data. The detailed criteria, presented in Table~\ref{tab:report-rubric}, were shared with students in advance to ensure transparency and alignment with expectations.

\begin{table}[ht]
\centering
\caption{Project report evaluation rubric}
\label{tab:report-rubric}
\renewcommand{\arraystretch}{1.3}
\begin{tabularx}{\textwidth}{>{\raggedright\arraybackslash}X c}
\toprule
\textbf{Criterion} & \textbf{Max Points} \\
\midrule
State-of-the-art implementation beyond class materials & 0.5 \\
Presentation and creativity in structure and design & 0.5 \\
Clarity of the best approach and results obtained & 1.0 \\
Experimental effort (number and diversity of experiments) & 2.0 \\
Analysis across individual patients & 2.0 \\
Conclusions on methods, tools, and limitations & 2.0 \\
Description of method construction and hyperparameters & 2.0 \\
\midrule
\textbf{Total} & \textbf{10.0} \\
\bottomrule
\end{tabularx}
\end{table}

\paragraph{Registration Accuracy Score (R Score).}
To isolate technical performance, the image registration outcome was independently evaluated using a composite metric. The R score combined three standard metrics in medical image alignment:

\begin{equation}
    R = 0.2 \times \text{TRE} + 0.3 \times \text{HD} + 0.5 \times (100 \times \left|\text{VS}\right|).
\end{equation}

\begin{itemize}
    \item $\text{TRE}$ (Target Registration Error) evaluates anatomical landmark alignment.
    \item $\text{HD}$ (Hausdorff Distance) captures outlier discrepancies in surface matching.
    \item $\text{VS}$ (Volume Similarity) measures volumetric agreement.
\end{itemize}

This component—worth 5\% of the final grade—served as a quantitative benchmark and encouraged students to optimize experimental configurations with rigor. The best-performing team received full credit, with the rest graded proportionally.

\paragraph{Oral Presentation Rubric.}
Each student delivered a brief oral presentation, which was assessed by both instructors and peers using a structured rubric. The criteria emphasized presentation structure, depth of understanding, and clarity of communication, supporting the development of individual responsibility and articulation within a collaborative learning environment. The complete evaluation matrix is provided in Appendix~\ref{appendix:presentation-matrix}.

\paragraph{Peer Evaluation.}
Peer assessment contributed 20\% of each student’s final grade, using a structured rubric aligned with oral presentation. This component encouraged individual responsibility, critical engagement with peers’ work, and reflective practice, all of which are key components of CSCL.

Together, these assessment tools formed a robust and multi-dimensional evaluation framework. By combining individual responsibility, team performance, and peer judgment, the approach not only reflected students’ technical competencies but also embedded collaborative and reflective learning within the grading structure.

\subsubsection{Collaborative and Peer Grading Rationale}

\paragraph{Peer Evaluation Design.} 
A structured peer grading system was implemented to assess individual contributions and to foster critical evaluation both within and between teams. The students were required to evaluate the oral presentations of all their classmates, excluding their own teammates, using a standardized rubric. The evaluations were anonymously submitted and contributed 20\% to each student’s final grade. Instructor assessments were conducted in parallel using the same rubric, ensuring triangulation of evaluation sources.

\paragraph{Pedagogical Rationale.} 
The peer evaluation process served both as a summative assessment tool and as a formative learning strategy. Grounded in CSCL principles, it promoted shared responsibility, peer-to-peer evaluation, and reflective thinking. Engaging students in evaluative judgment deepened their understanding of quality criteria, enhanced critical thinking, and supported the co-construction of knowledge. In a project-based learning environment, peer assessment positioned learners as active contributors to both knowledge creation and quality assurance, aligning with broader educational goals in collaborative STEM education. 

\subsection{Student Feedback and Perception Analysis}
To assess the perceived impact of the pedagogical interventions, particularly the integration of W\&B, other collaborative tools, and peer grading, a structured Google Form was administered at the end of each academic year. The survey gathered student perspectives on several dimensions, including engagement, tool usability, collaborative dynamics, and the fairness and usefulness of peer evaluation. 

Questions combined quantitative measures such as Likert-scale ratings on usefulness and satisfaction with open-ended qualitative responses. This mixed-methods design provided a detailed view of how the methodology influenced students’ learning experiences, technical development, and teamwork.

The results informed a longitudinal comparison across the two cohorts (2023/2024 and 2024/2025), enabling reflection on both the stability and evolution of student perceptions under the same instructional design. The insights informed course improvements and contributed empirical evidence to the wider discussion on CSCL-based practices in advanced STEM education.

\section{Results}\label{sec:results}

\subsection{Survey-Based Evaluation of Student Experience}
Surveys were conducted in the 2023/2024 and 2024/2025 academic years to examine the effectiveness and reception of the revised pedagogical model. They captured student perceptions of tool effectiveness, collaborative learning, evaluation clarity, and skill development.

\subsubsection{Cohort 2023/2024: First-Year Implementation}
\begin{figure}
    \centering
    \includegraphics[width=\textwidth]{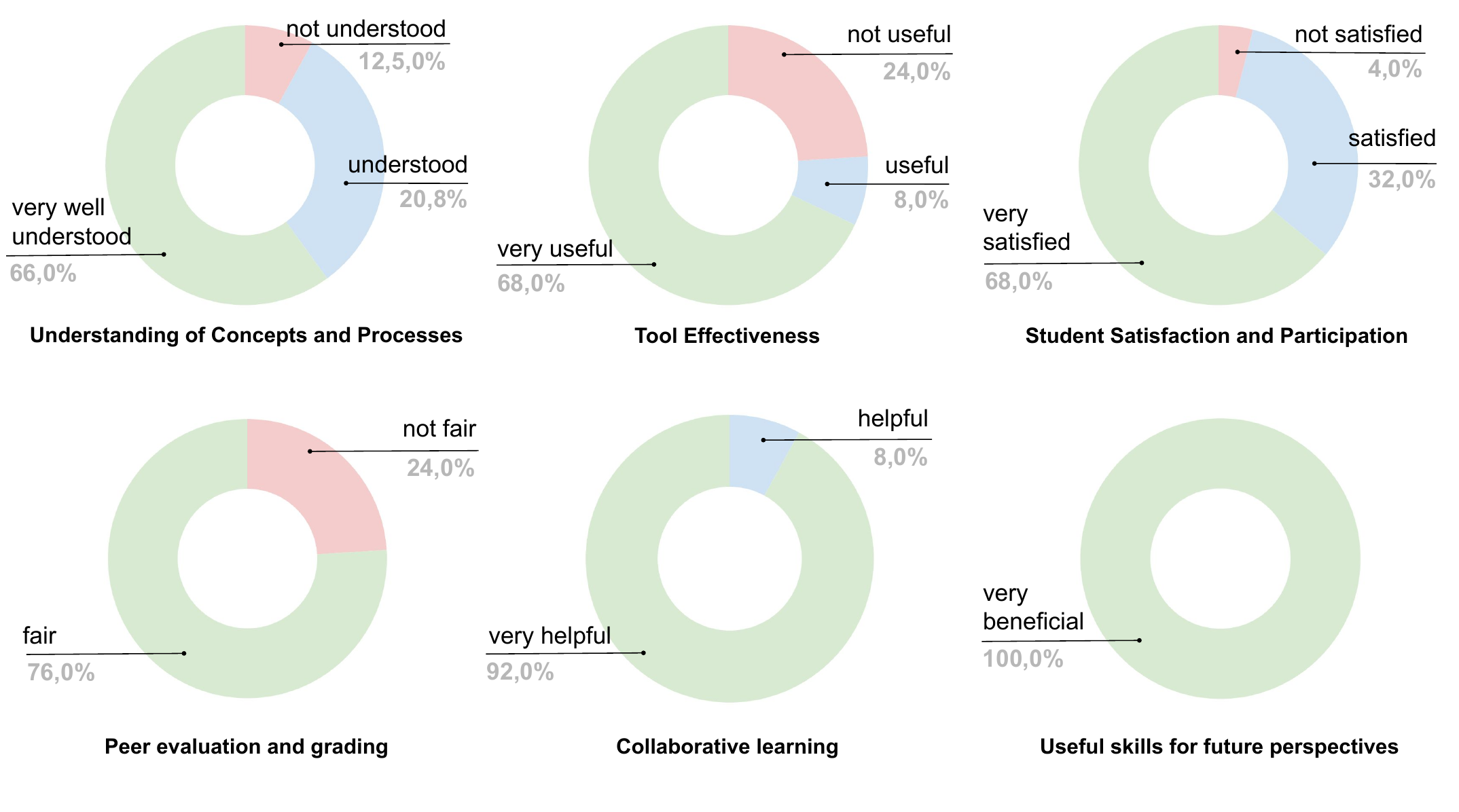}
    \caption{Summary of survey responses from the 2023/2024 cohort evaluating key aspects of the project-based learning experience. Each donut chart illustrates the proportion of students reporting different levels of agreement or satisfaction across six dimensions: understanding of concepts and processes, tool effectiveness, overall satisfaction, fairness of peer evaluation, collaborative learning, and usefulness of acquired skills. Green indicates the most positive responses (e.g., "very well understood," "very useful"), blue represents intermediate or neutral feedback (e.g., "understood," "useful"), and red reflects less favorable responses (e.g., "not useful," "not fair"). The highly positive trend across all categories indicates strong endorsement of the tool-integrated collaborative learning model.}
    \label{fig:survey-results}
\end{figure}

At the end of the 2023/2024 course, a structured survey was distributed to assess the educational impact of the newly implemented CSCL-based methodology. It focused on six dimensions: conceptual understanding, tool effectiveness, student satisfaction, grading clarity, collaborative learning, and perceived skill development. Responses were recorded on a five-point Likert scale from 1 (strongly disagree) to 5 (strongly agree). A summary of the aggregated results is shown in Figure~\ref{fig:survey-results}.

\paragraph{Conceptual Understanding.}  
Students reported a strong grasp of the core principles behind deformable image registration. Most rated their understanding at level 4 or higher, indicating that the integration of applied coding tasks with theoretical content effectively supported knowledge acquisition.

\paragraph{Tool Effectiveness.}  
W\&B was highly rated for its role in experiment tracking, visualization, and report generation. Most of the students assigned it a 4 or 5, suggesting that its integration meaningfully enhanced their ability to document and interpret experimental results within a collaborative context.

\paragraph{Student Satisfaction and Engagement.}  
Overall satisfaction with participation in the project was high. Students appreciated the autonomy offered by the project framework and felt that their contributions were valued. This reflects positively on the project’s structure and alignment with authentic learning principles.

\paragraph{Grading Transparency and Evaluation Tools.}  
The presentation evaluation matrix was perceived as especially helpful, with most students agreeing that it clarified expectations and improved their preparation. This underscores the importance of transparent, rubric-driven assessment in supporting self-regulated learning.

\paragraph{Collaborative Learning and Skill Development.}  
Students consistently reported that the project supported collaboration and the development of technical and transversal skills. These included team coordination, problem-solving, and effective communication, all of which are central competencies to CSCL environments and highly transferable to professional practice.

\paragraph{Preparation for Future Practice.}  
Notably, students highlighted W\&B not only as useful within the course but also as a valuable tool for future research and industry contexts. This suggests that the project successfully bridged the gap between classroom learning and real-world applications.

Together, these results indicate that the first implementation of the collaborative, tool-supported methodology achieved its pedagogical goals. Student feedback confirms the value of combining open-ended project work with structured support tools and assessment frameworks. Areas for refinement, such as early onboarding with W\&B or clearer scaffolding in the project timeline, were identified and informed the design of the following cohort's experience.

\subsection{Survey-Based Evaluation of the Student Experience (2024/2025)}

\begin{figure}
    \centering
    \includegraphics[width=\textwidth]{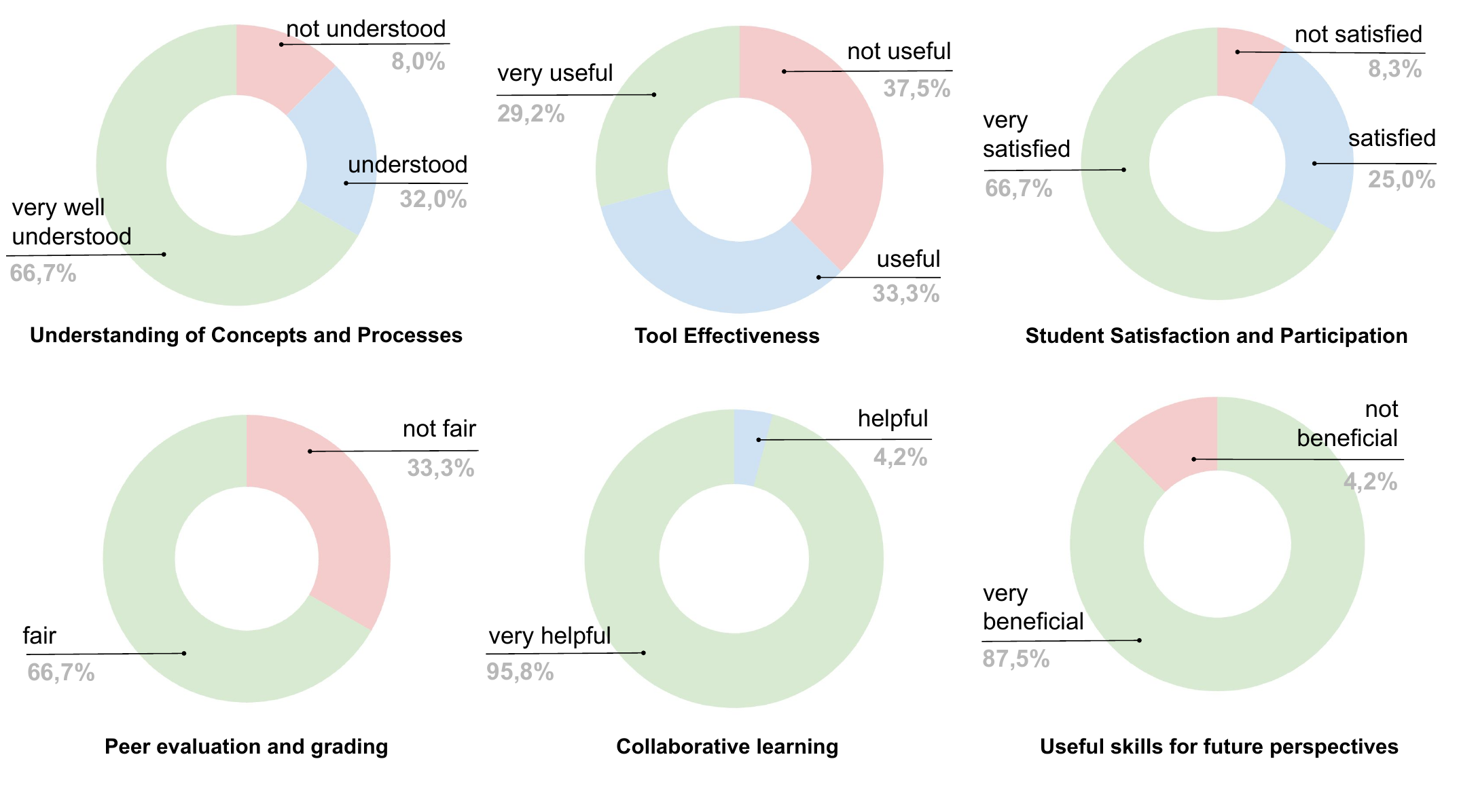}
    \caption{Summary of survey responses from the 2024/2025 cohort regarding their project experience. Each donut chart visualizes student feedback across six key areas: conceptual understanding, tool effectiveness, satisfaction, peer evaluation fairness, collaborative learning, and skill development. Colors follow the same convention: green (most positive), blue (neutral), and red (less favorable). While responses remained predominantly positive, slight decreases in perceived fairness and tool usefulness were observed compared to the previous cohort, potentially reflecting shifting student expectations or external workload pressures.}
    \label{fig:survey-results-2425}
\end{figure}

To assess the pedagogical impact of the 2024/2025 implementation, we repeated the survey from the previous year. The results, shown in Figure~\ref{fig:survey-results-2425}, provide a comparison of the ongoing integration of collaborative strategies and technical tools in biomedical image processing education.

\paragraph{Understanding of Concepts and Processes.}
The majority of students (66.7\%) reported a strong grasp of the core concepts, with an additional 32.0\% indicating satisfactory understanding. Only 8.0\% reported difficulty. These results reflect a robust assimilation of deformable registration principles, with minimal conceptual barriers.

\paragraph{Tool Effectiveness.}
Perceptions of W\&B and Google Colab were more mixed than in the previous year. While 29.2\% rated the tools as very useful and 33.3\% as useful, a notable 37.5\% perceived them as not useful. This suggests possible challenges in tool onboarding or alignment with team workflows, which warrants additional instructional scaffolding.

\paragraph{Satisfaction and Participation.}
Two-thirds of the cohort (66.7\%) expressed high satisfaction with their engagement in the project, and another 25.0\% were moderately satisfied. This indicates sustained student motivation and a perceived sense of ownership over the learning experience.

\paragraph{Peer Evaluation and Grading.}
66.7\% of respondents considered the peer evaluation system fair, while 33.3\% found it unfair. This split highlights both the potential and limitations of student-led assessment mechanisms and emphasizes the need for continued calibration and transparency in peer grading design.

\paragraph{Collaborative Learning.}
A striking 95.8\% of students rated the project as very helpful for collaborative learning, affirming the efficacy of the CSCL-informed instructional design. Only one student (4.2\%) indicated limited benefit.

\paragraph{Skill Development for Future Contexts.}
The project was overwhelmingly recognized as a beneficial skill-building experience, with 87.5\% of students rating it as very beneficial for future academic or professional work. This affirms the long-term relevance of the project’s design in cultivating transferable competencies beyond course-specific objectives.

Overall, the results validate the teaching model introduced in the prior year. While positive trends in conceptual learning and collaboration persist, shifts in perceived tool usefulness and peer assessment fairness invite reflection. These insights will guide targeted refinements in tool onboarding and peer evaluation calibration for future iterations.

\subsection{Grade Distribution Trends Across Three Cohorts}
To assess the pedagogical impact of the redesigned evaluation framework, we analyzed the grade distributions across three consecutive academic years (2022/2023, 2023/2024, and 2024/2025) of the Biomedical Image Processing course. These distributions provide insight into how the shift toward a collaborative, rubric-driven model influenced both assessment fairness and performance differentiation.

Prior to the implementation of the revised methodology, the 2022/2023 cohort exhibited a grade distribution heavily skewed toward the upper range of the scale. Most students scored between 9 and 10 out of 10, with a minimum grade of 7.5, indicating limited dispersion and raising concerns about grade inflation and insufficient differentiation in learning outcomes.

In contrast, the 2023/2024 cohort, assessed using the redesigned, rubric-based framework integrating collaborative and peer-assessment elements, showed a more balanced distribution of scores (5.8–8.9). This indicated a more detailed appraisal of individual competencies, including programming, communication, critical analysis, and teamwork. This pattern was further consolidated in the 2024/2025 cohort, which followed the same methodology. The third-year distribution retained the bell-shaped form, with slightly increased dispersion (6.55–9.79) and clear differentiation of grades across a wide range of performance.

To quantify the increase in distributional fairness and variability, we computed the Shannon entropy ($H$) for each cohort’s grade distribution. In this context, entropy is a statistical measure of diversity in a distribution. Originally developed in information theory, Shannon entropy captures how evenly outcomes (in this case, grades) are spread across categories. 

A higher entropy value reflects a more diverse and informative grade distribution, suggesting that the evaluation framework distinguishes more effectively between different levels of student performance. In contrast, lower entropy typically indicates grade compression or inflation, where a majority of students receive similar marks regardless of individual learning outcomes.

The entropy of each grade distribution was calculated as follows:

\begin{equation}
H = - \sum_{i=1}^{n} p_i \log_2 p_i ,
\end{equation}

where $p_i$ represents the proportion of students who receive grade $i$. The entropy values for the three cohorts were as follows:
\begin{itemize}
    \item 2022/2023: $H = 0.87$
    \item 2023/2024: $H = 1.55$
    \item 2024/2025: $H = 1.81$
\end{itemize}

The marked increase in entropy over the years confirms that the updated evaluation design produces more differentiated and equitable student outcomes. As described in Section~\ref{subsec:evaluation}, the multi-component grading structure, which includes peer-assessed presentations, structured reports, and objective R-scores, provides a robust framework for evaluating diverse student skills and reducing grade compression.

\begin{figure}
    \centering
    \includegraphics[width=\textwidth]{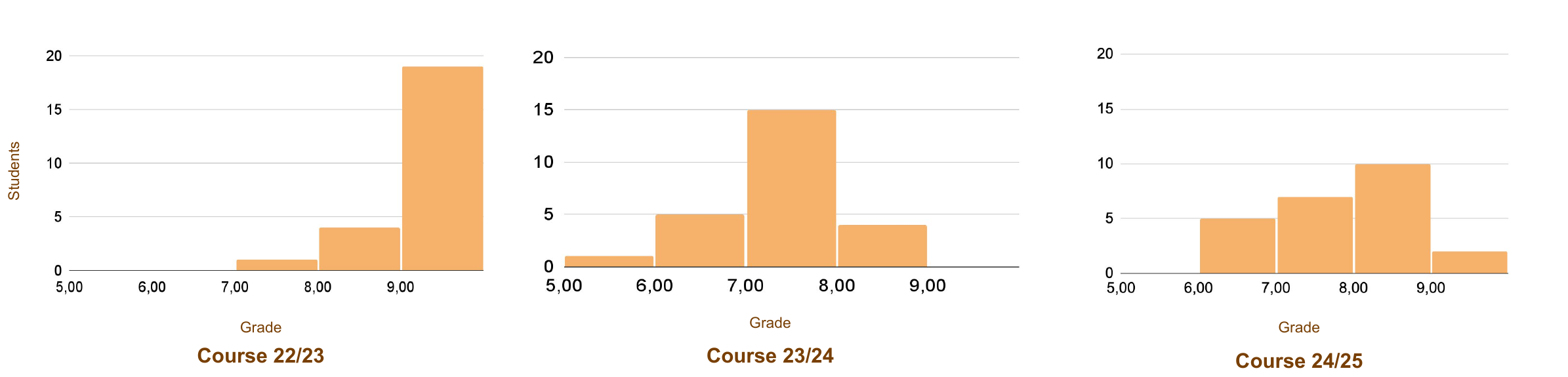}
    \caption{Grade distributions across three academic years (2022/2023 to 2024/2025). The transition to a rubric-based, collaborative evaluation framework is associated with a broader and more balanced distribution of student outcomes. Shannon entropy values for each year are: 2022/23-0.87, 2023/24-1.55, and 2024/25-1.81, indicating a progressive increase in distributional diversity.}
    \label{fig:grade-distribution}
\end{figure}

\subsection{Evaluation Transparency and Structural Fairness}

To complement the analysis of grade distributions, we examined how the structural design of the evaluation framework supported objectivity and transparency. Rather than relying solely on instructor discretion, the project grade was derived from a multi-component system combining rubric-based instructor assessment, peer evaluation, and quantitative performance metrics such as the R-score. This structure aimed to reduce subjectivity, align student expectations, and diversify sources of feedback.

Survey data from two consecutive cohorts supported the effectiveness of this approach. Students consistently rated the evaluation matrix as highly useful in guiding their preparation for presentations, suggesting that clear assessment criteria enhanced their sense of control and focus during the project. Opinions on peer grading were more divided. Some students questioned its fairness, while others recognized its role in promoting critical engagement and mutual responsibility. This mix of views reflects a well-documented tension in the peer assessment literature and underscores the importance of structured rubrics to support their implementation. 

Objectively, the inclusion of the R-score, a composite metric derived from landmark and volumetric registration accuracy, anchored part of the final grade in reproducible experimental outcomes. This quantitative foundation complemented the more subjective elements of the evaluation, contributing to a robust and multidimensional assessment framework. The triangulation of assessment sources (instructors, peers, and algorithmic results) reflects best practices in educational evaluation and aligns with CSCL principles, which emphasize shared responsibility and evaluative co-agency.

In summary, although students' views on peer grading varied, the underlying structure of the evaluation system demonstrably increased clarity, fairness, and alignment with the learning objectives. These design elements contributed to the observed normalization of grade distributions and offer a replicable model for collaborative, skills-integrative assessment in STEM education.

\section{Discussion}\label{sec:disc}

This study examined the implementation of a redesigned pedagogical strategy for a biomedical image processing project in three academic cohorts. The combined use of modern collaborative tools (W\&B, Google Colab), structured evaluation (rubrics and peer grading), and authentic project-based learning formed the foundation of an instructional model grounded in CSCL principles and digital literacy. The results over multiple years suggest both pedagogical robustness and scalable applicability to data-intensive STEM domains.

Consistent with prior research on the use of collaborative environments to improve learning outcomes (\cite{KUROKI2021100225, baptista2021using, canesche2021google}), our findings show that Google Colab facilitated real-time teamwork and transparent co-development. The introduction of W\&B, though initially met with reluctance, became central to the course workflow. In line with \cite{WU2024, WEN2022101463, ALKHALED2023100030}, our experience confirmed the importance of supporting tool adoption through in-class demonstrations and targeted guides to increase student acceptance and reduce resistance to new technologies.  

A key outcome of this methodological change was the increased traceability and detail of student contributions, made possible through W\&B’s logging features. These not only facilitated formative supervision by instructors but also equitable summative assessment. Despite the preference for anonymity reported in other contexts (\cite{VELAMAZAN2023104848}), the visible attribution of W\&B runs encouraged individual responsibility within collaborative tasks.

Peer grading and structured rubrics further contributed to evaluation transparency. Following the evidence presented in \cite{brookhart2018appropriate}, rubrics were distributed prior to presentations, providing clear expectations and fostering more focused student preparation. Over the years, students consistently valued this transparency, as reflected in survey responses and informal feedback. However, views on peer assessment were mixed: many students found it fair and constructive, while others raised concerns about potential bias, particularly the influence of interpersonal relationships. These concerns align with broader challenges in collaborative assessment and suggest the need for additional training or hybrid grading models in future iterations. 

Quantitative analysis of grade distributions over three academic years further supports the effectiveness of the new methodology. The 2022/2023 cohort, evaluated under the traditional model, showed a compressed grade range (mean = 9.27, SD = 0.33), raising concerns about grade inflation and insufficient differentiation. In contrast, the subsequent cohorts (2023/2024 and 2024/2025) presented broader and more symmetric distributions (SD = 0.70 and 0.90, respectively), indicating greater sensitivity to individual performance. Importantly, Shannon entropy, used as a measure of distributional fairness and differentiation, increased from 0.87 (2022/2023) to over 1.5 in the subsequent years. This metric supports the conclusion that the revised framework improves the discriminatory power of assessment and better reflects individual learning outcomes. 

In line with recent studies (\cite{qureshi2023factors, wakeman2022fair}), this evolution toward rubric-guided, data-supported evaluation promotes both equity and rigor. Nonetheless, some logistical challenges remain. The linear editing constraints of W\&B create difficulties for simultaneous collaboration, and the quality of report submissions varied considerably in creativity and depth, suggesting the need for clearer templates or formatting guidelines. These limitations mirror broader issues in project-based learning, where balancing structure and autonomy remains an ongoing pedagogical challenge (\cite{SANTOS2015354}).

The sustained success of this methodology over three years also demonstrates its potential for transferability. The initial trial in the AI in Health course, part of a newly launched Applied AI Master’s program, served as a testing ground for experimentation. Although small in scale, this early experience confirmed the feasibility of using these tools for data-intensive, collaborative learning contexts (\cite{lefevre2022modelops}). Since then, the approach has proved scalable and sustainable in the Biomedical Image Processing course, suggesting its value as a model for other STEM curricula seeking to integrate CSCL principles with practical toolchains and multimodal evaluation. 

While the findings demonstrate the effectiveness and scalability of the proposed methodology, several limitations should be acknowledged. First, the study was conducted within a single institution and academic discipline, which may limit generalizability. However, the structured and well-documented nature of the intervention facilitates replication in related STEM fields. Second, although peer evaluation offers opportunities for critical reflection and shared responsibility, it can be influenced by interpersonal dynamics or inconsistent grading standards. This risk was mitigated through standardized rubrics and triangulation with instructor evaluations and objective performance metrics. Third, varying levels of familiarity with tools such as W\&B may have affected engagement and perceived usefulness. To address this, onboarding and support materials were provided, with refinements based on student feedback. Finally, as with any longitudinal design, cohort-specific variables such as group dynamics, instructor style, or competing academic deadlines could have influenced student engagement or outcomes. Nevertheless, consistent positive trends across two independent cohorts and the inclusion of a pre-intervention baseline strengthen the robustness of the conclusions.

These limitations reflect the realities of authentic educational environments and, rather than weakening the findings, highlight their relevance and applicability to real-world teaching contexts.

\section{Conclusions}\label{sec:conclusion}

This study reports on a multi-year implementation of a redesigned pedagogical framework in the Biomedical Image Processing course, integrating collaborative learning strategies, structured evaluation, and digital tools into project-based STEM education. The approach combined cloud-based coding environments, experiment-tracking platforms, and rubric-guided assessment to promote both disciplinary expertise and transferable skills such as teamwork, critical analysis, and communication.

Across two implementation years, analysis of grade distributions and entropy measures indicated more equitable and differentiated assessments compared with the pre-intervention baseline. Student feedback highlighted the value of transparent evaluation criteria and the benefits of collaborative tool use, while also identifying peer assessment as both a strength and an area for refinement.

Originally piloted in a smaller-scale AI in Health course, the methodology proved adaptable and sustainable in a more advanced, data-intensive setting. Its documented processes, measurable outcomes, and positive reception suggest potential for replication in other STEM contexts that require collaborative problem-solving and authentic project work.

Future research should examine the longer-term impact of these practices on student development and career readiness, as well as explore adaptations for interdisciplinary and cross-institutional projects. Continued refinement of peer evaluation processes and tool integration will be important to enhance fairness, efficiency, and learner engagement. 

Overall, this work offers a transferable model for combining collaborative learning, transparent assessment, and modern digital platforms in technical higher education, contributing both empirical evidence and practical guidance for educators designing engaging, equitable, and skill-oriented learning environments.

\newpage

\section{Appendix: Evaluation Rubric for Student Presentations}\label{appendix:presentation-matrix}
\begin{longtable}{|p{0.15\linewidth}|p{0.55\linewidth}|p{0.2\linewidth}|}
\caption{Evaluation Matrix for Student Presentations.}
\label{tab:presentation_matrix} \\
\hline
\textbf{Criteria} & \textbf{Description} & \textbf{Mark} \\
\endfirsthead
\hline
\endhead
\hline
\endfoot
\endlastfoot
\hline
\textbf{Structure \qquad (1 point)} & 
\begin{itemize}
    \item The speaker greets the audience and presents themselves.
    \item Starts with a brief description of the presentation's structure (index).
    \item Develops the body of the presentation, signposting each part's conclusion and the next part's start.
    \item Concludes with a summary of the main ideas.
\end{itemize} & 
\begin{itemize}
    \item 1 point if all criteria are met
    \item 0.5 point if one criterion is missing
    \item 0 points if two or more criteria are missing
\end{itemize} \\
\hline
\textbf{Understanding of the registration method \qquad \qquad (3 points)} & 
\begin{itemize}
    \item The student understands and accurately explains the image registration method used.
    \item They provide a convincing rationale for the chosen approach.
\end{itemize} & 
\begin{itemize}
    \item 3 to 0 points, depending on the clarity and accuracy of explanation
\end{itemize} \\
\hline
\textbf{Content and precision \qquad \qquad (3 points)} & 
\begin{itemize}
    \item The student offers all necessary details to replicate the method used.
    \item The code must function automatically.
    \item The approach is correct, and the presentation provides a comprehensive overview.
    \item Explanations are clear, concise, and free of superfluous content.
    \item Appropriate terminology and formality level are maintained throughout.
\end{itemize} & 
\begin{itemize}
    \item 3 points if all criteria are met
    \item 2 points if one criterion is missing
    \item 1 point if two criteria are missing
    \item 0 points if three or more criteria are missing
\end{itemize} \\
\hline
\textbf{Delivery \qquad \qquad (1 point)} & 
\begin{itemize}
    \item Appropriate volume and pacing are maintained.
    \item The tone is varied to emphasize key points.
    \item Speech is clear, avoiding filler words and colloquialisms.
\end{itemize} & 
\begin{itemize}
    \item 1 point if all criteria are met
    \item 0.5 points if one criterion is missing
    \item 0 points if two or more criteria are missing
\end{itemize} \\
\hline
\textbf{Slides \qquad \qquad (1 point)} & 
\begin{itemize}
    \item Slides are free from errors and text-heavy sections.
    \item Font size is adequate (minimum 24).
    \item Images used are clear and support the content explained.
\end{itemize} & 
\begin{itemize}
    \item 1 point if all criteria are met
    \item 0.5 points if one criterion is missing
    \item 0 points if two or more criteria are missing
\end{itemize} \\
\hline
\textbf{Gestures and body expression (0.5 point)} & 
\begin{itemize}
    \item The speaker maintains good posture and uses gestures effectively.
    \item Eye contact is established with the entire audience.
\end{itemize} & 
\begin{itemize}
    \item 0.5 point if all criteria are met
    \item 0.25 points if one criterion is missing
    \item 0 points if two or more criteria are missing
\end{itemize} \\
\hline
\textbf{Time \qquad \qquad(0.5 point)} & 
\begin{itemize}
    \item The presentation is adjusted to the established time and is distributed equally among team members.
    \item For an ‘ideal’ duration of 10 minutes:
    \begin{itemize}
        \item 0.5 point if it takes between 9 and 11 minutes
        \item 0.25 points if it takes between 8 and 12 minutes
        \item 0 points if it takes less than 8 minutes or more than 12 minutes
    \end{itemize}
\end{itemize} & \\
\hline
\end{longtable}

\section*{Acknowledgments}\label{sec:ack}

Special thanks to Jaume Ramis Bibiloni for generously sharing his evaluation matrix, which informed the design of our assessment rubrics. We also gratefully acknowledge Alejandro Guerrero-López and María Martínez-García, whose early implementation of Weights \& Biases in the Applied AI master's program inspired the pedagogical innovations explored in this study. Their contributions were instrumental in shaping the methodology adopted in the Biomedical Image Processing course.

\printcredits

\bibliographystyle{cas-model2-names}

\bibliography{main-cas-sc-template}

\bio{}
\endbio

\endbio

\end{document}